\documentclass[letterpaper, 10 pt, conference]{ieeeconf}  

\IEEEoverridecommandlockouts                              

\usepackage{amsmath} 
\usepackage{amssymb}  
\usepackage{algorithmic}
\usepackage{algorithm}
\usepackage{bm}
\usepackage{color}
\usepackage{graphicx}
\usepackage{cite}

\title{\LARGE \bf
Online Adaptation of Learned Vehicle Dynamics Model with Meta-Learning Approach
}

\author{Yuki Tsuchiya$^{1}$, Thomas Balch$^{2}$, Paul Drews$^{3}$ and Guy Rosman$^{2}$
\thanks{$^{1}$Yuki Tsuchiya was with Toyota Research Institute, Cambridge, MA, USA. He is now with Toyota Motor Corporation, Toyota City, Aichi, Japan
        {\tt\small yuki\_tsuchiya\_aa@mail.toyota.co.jp}}%
\thanks{$^{2}$Thomas Balch and Guy Rosman are with Toyota Research Institute, Cambridge, MA, USA
        {\tt\small thomas.balch@tri.global, guy.rosman@tri.global}}%
\thanks{$^{3}$Paul Drews was with Toyota Research Institute, Cambridge, MA, USA
        {\tt\small paul.drews@gmail.com}}%
}

\begin{document}

\maketitle
\thispagestyle{empty}
\pagestyle{empty}

\begin{abstract}
We represent a vehicle dynamics model for autonomous driving near the limits of handling via a multi-layer neural network. Online adaptation is desirable in order to address unseen environments. However, the model needs to adapt to new environments without forgetting previously encountered ones. In this study, we apply Continual-MAML to overcome this difficulty. It enables the model to adapt to the previously encountered environments quickly and efficiently by starting updates from optimized initial parameters. We evaluate the impact of online model adaptation with respect to inference performance and impact on control performance of a model predictive path integral (MPPI) controller using the TRIKart platform. The neural network was pre-trained using driving data collected in our test environment, and experiments for online adaptation were executed on multiple different road conditions not contained in the training data. Empirical results show that the model using Continual-MAML outperforms the fixed model and the model using gradient descent in test set loss and online tracking performance of MPPI.
\end{abstract}

\section{INTRODUCTION}
Autonomous vehicles have been gaining attention since they have the potential to outperform humans with regards to safety, traffic congestion relief, and alleviation of motion sickness for passengers. Recently, some companies have started to deploy autonomous vehicles in the real world\cite{c1,c2}. However, their deployment is limited to specific regions and conditions. Thus, there is a need to expand the operating envelope of autonomous vehicles so that they can travel safely in any driving situation and condition.\par
To improve safety and confidence in emergency scenarios, for example, the appearance of an unexpected obstacle, high performance control near the limits of handling is required. Autonomous racing is a good proxy problem for limit handling in autonomous vehicles\cite{c3,c4,c5,c6,c7,c8}. Some cutting-edge research even tackles stable control while in a drifting state in order to maximize vehicle potential\cite{c9,c10}. \par
For local planning and control of autonomous vehicles, model-based approaches are the most common. To achieve high performance, an accurate model with fast inference time is required. Many model-based approaches represent vehicle dynamics via a physics-based model, as it can provide  reasonable vehicle behavior and is easy to analyze\cite{c3,c7,c9}. Although the physics-based approach is a powerful and proven method, it becomes complex or fails to model highly nonlinear behavior near the limits of handling. On the other hand, learned models such as neural network models have the potential to capture nonlinear features without a significant increase in computational cost\cite{c10,c11,c12}. For these reasons, we adapt a neural network as a learned dynamics model to surpass the performance of the physics-based models.\par
Learned model performance strongly depends on the training data set. In other words, applying a fixed model to an environment that is not included in the training set can cause poor performance. For autonomous vehicles, there are a lot of elements that vary over time, such as passenger and load weights, tire conditions, and road conditions. Tires can be changed by the individual user to cope with winter road conditions or to replace old tires with new ones. In\cite{c5,c6}, a learned model is developed for two different road conditions, high- and low-friction roads, by integrating historical data into the model input. However, their training set contains both high- and low-friction road data and it is difficult to guarantee the performance for all possible scenarios. There are immeasurable environments in the real world and it is impractical to infer a model from all of them in advance. Therefore, online model adaptation is a powerful alternative to adapt to possible environments that are not contained in the training set.\par
When driving in real-time, the vehicle will encounter unseen environments as mentioned above, as well as environments encountered in the past. Therefore, an online adaptation algorithm needs to not only adapt to a new environment, but also use past information to quickly re-adapt to a previously encountered environment. So far, few methods have stood out in efficient sequential learning of vehicle dynamics with learned models, and research in this field is important for developing vehicle control robust to the constant changes in the real world. In this paper, we address the online adaptation problem in learned vehicle dynamics models for model-based control. We apply Continual-MAML\cite{c13} to our learned vehicle dynamics model in order to achieve high control performance when conditions revert to a previous one, while maintaining the ability to adapt to unforeseen conditions. Continual-MAML is an extension of the well-known meta-learning method, MAML, designed to address scenarios where new and previous tasks emerge. It facilitates efficient learning in such situations. When the vehicle encounters a previously observed road condition, the model can adapt efficiently by starting parameter updates from an optimal initial value based on prior observations. As a result, model-based control shows higher performance.\par
To evaluate the impact of model improvements on a model-based controller, we implement a model predictive path integral (MPPI) controller\cite{c14,c15,c16,c17,c18}. This controller is derived from model predictive control (MPC) and utilizes a sampling-based approach for optimization. MPPI is a powerful control algorithm because it naturally takes the system nonlinearity into account.\par
We conducted our experiments using a  one-tenth scale-car\cite{c19}. The scale-car makes it possible to capture several realistic elements which are difficult to model in a simulation and it is much simpler to ensure the safety of the experiment versus using a full-scale car. We compare inference and control performances for three models, namely, a fixed model, an online adaptive model with gradient descent and an online adaptive model with Continual-MAML in a scenario where two different road surfaces appear in turn. Driving data on those surfaces are not included in the training data in order to see the adaptive performance of the models and each model uses the same neural network structure. We show that online adaptation with Continual-MAML outperforms the fixed model and adaptive model with gradient descent in terms of both inference and control performance. We focus on differences in road conditions. However, our approach can be extended to other unforeseen scenarios, such as variable loads.\par
To summarize, our contributions include: (1) applying the Continual-MAML algorithm to a learned vehicle dynamics model and providing a description of its application. (2) contrasting with alternative approaches and showing the Continual-MAML outperforms a fixed model and an adaptive model with gradient descent for both inference and control performance with model-based control on a real platform.
\section{RELATED WORK}
Our work relates to vehicle dynamics modeling for planning and control, to its use in online adaptation, and to meta-learning approaches.
\subsection{Vehicle Dynamics Modeling}
For model-based control in the autonomous vehicle field, physics-based approaches are the popular choice because they are  easy to analyze and a simple physics-based model is usually feasible and accurate enough for ordinary driving situations. In\cite{c20} it is shown that a kinematic bicycle model is as accurate as a dynamic model which takes into consideration slip, when lateral acceleration is not large. \cite{c21} provides a threshold for lateral acceleration so that the kinematic bicycle model for motion planning is reliable. Even in emergency scenarios such as obstacle avoidance or driving near the limits of handling, like drifting, autonomous driving has been achieved using nonlinear model predictive control (NMPC) with a combination of a bicycle model and a brush tire model\cite{c9,c22}.\par
An alternative method to model the vehicle dynamics is a learning-based approach. Recently, learned models for vehicle dynamics have attracted attention because they have the potential to capture the nonlinear aspects of the vehicle dynamics without a significant increase in computational cost. In \cite{c5,c6}, the learned model is used as a vehicle dynamics model for path tracking control on high- and low-friction road surfaces and they outperform the control performance of a physics-based model consisting of a bicycle model and a brush tire model. Parameters of the learned model are learned from both road surfaces and parameters of the physics-based model are tuned to each. \cite{c11} uses a simple neural network model to estimate slip angle and realizes high performance without high computation cost and additional sensors. \cite{c12} uses a recurrent neural network to model steering system dynamics and combine it with a physics-based vehicle model for autonomous driving. \cite{c10} proposes a learned tire model which satisfies physical assumptions for use with an autonomous drift controller. Their learned model has higher prediction accuracy than Fiala brush and Magic Formula tire models.
\subsection{Online Adaptation}
A model-based controller requires a high-fidelity model. In the real world, however, it is possible for vehicles to encounter new environments that were not considered when the model was designed. Online adaptation addresses this shortcoming.\par
For online adaptation of physics-based models, several studies focus on the online identification of the friction between the road and the tire because it is the critical parameter affecting vehicle behavior and it varies based on the tire state and road conditions. \cite{c3} estimates the friction parameter for autonomous driving using an unscented Kalman filter. \cite{c7} demonstrates that an autonomous vehicle at the friction limit can track a circular path with an online friction estimator. Alternatively, \cite{c4} doesn't only focus on the friction parameter.  Their approach estimates the error of the physics-based model in the state equation with Gaussian Process regression to enhance the performance of autonomous driving with a MPC. \cite{c23} adapts a similar approach that expresses a nominal model with a kinematic based model and also compensates  differences between it and dynamics model by using a learned tire model.\par
For online adaptation of learned models, the primary focus is  adapting to new environments that are not included in the training set. For learned models, sequential learning is required in order to do this. \cite{c24} develops an online parameter update algorithm using gradient descent with momentum. Until the learned model outperforms the physics-based model, this approach uses the physics-based model and then, when the learned model achieves a higher performance, the model is switched to the learned model. A major issue of online adaptation for a neural network model is catastrophic forgetting, which is a phenomenon where the model forgets old data when adapting to new data. To avoid it, \cite{c25} incorporates locally weighted projection regression into the update scheme. \cite{c26} incorporates the idea of latent vectors. It infers the local road surface condition from history states and actions, acoustic spectrogram readings and RGB images. The learned dynamics model using latent vectors provides shorter lap times and safer driving than the model not using latent vectors on different surface conditions.
\subsection{Meta-Learning}
To make a model quickly adapt to a new environment, one key factor is the efficiency of the update algorithm. Meta-learning is also called learning to learn because it improves the learning algorithm itself. It aims to make agents which quickly adapt to a new task. Model-Agnostic Meta-Learning (MAML) is the meta-learning method which can quickly solve new tasks by learning initial model parameters \cite{c27}. By starting the parameter update cycle from learned initial values, the model can quickly adapt to each task it encounters. MAML assumes that tasks are available together as a batch and it is not capable of ingesting sequential tasks. Therefore, there is increasing interest in applying MAML to an online scheme where data is measured sequentially. Follow The Meta Leader (FTML) \cite{c28} applies MAML to an online framework. FTML can update initial parameters even if the data set of an unseen task is measured sequentially.  However FTML needs to store all the observed data and this feature is not desirable for practical problems. Other research applies a meta-learning framework to the situation where data arrives incrementally without segmenting it into discrete tasks\cite{c29}. However, \cite{c29} tests its continuous learning performance on similar data to that from pre-training, and performance on new tasks is not guaranteed.  \cite{c13} focuses on the continuous-learning scenario called online fast adaptation and knowledge accumulation (OSAKA) where new and previous tasks appear. In order to cope with this scenario i.e. appearance of new tasks while adapting to previous tasks quickly once they reoccur, Continual-MAML is developed as an extension of the original MAML algorithm, but with the ability to run online.
\section{OUR APPROACH}

\subsection{Vehicle Dynamics Modeling}\label{3A}
To capture nonlinear characteristics of vehicle dynamics near the limits of handling without a significant increase in calculation cost, we utilize a multi-layer neural network to model the vehicle dynamics. To model the vehicle dynamics properly, we adapt the neural network model framework from\cite{c18}. In \cite{c18}, authors show that their multilayer neural network significantly outperforms their basis function derived from physics perspective.\par
We define the state variable vector as:
\begin{equation}
{\bm x}=
\begin{bmatrix}
\phi, & v_{\rm x}, & v_{\rm y}, & r
\end{bmatrix}^\mathsf{T}\label{eq1}
\end{equation}
\noindent
where $\phi$ is roll angle, $v_{\rm x}$ and $v_{\rm y}$ are $x$ and $y$ velocity in the local coordinate frame and $r$ is yaw rate. Then, the state equation is written as:
\begin{equation}
{\bm x}(t+1)={\bm x}(t)+{\bm f}({\bm x}(t),{\bm v}(t))\Delta t\label{eq2}
\end{equation}
\noindent
where $\Delta t$ is a sampling period and $v(t)$ is the system input vector:
\begin{equation}
{\bm v}(t)=
\begin{bmatrix}
u_1(t)+\epsilon_1\\
u_2(t)+\epsilon_2
\end{bmatrix}\label{eq3}
\end{equation}
\noindent
$u_1$ and $u_2$ are the steering and acceleration input, respectively, and they are randomly perturbed by gaussian noise, $\epsilon$. In \eqref{eq2},  $f(\cdot)$ is a fully connected multilayer neural network model with a hyperbolic tangent activation function. It has two hidden layers, each with 32 neurons. The model is trained using an Adam optimizer. It minimizes prediction loss for MPPI over 100 time steps. In our environment, training data is decimated for the sampling period to be 20 (ms) in length. Hence, the model is trained to optimize over a 2 (s) interval. The loss is the sum of the squared error of the X-position $X$, the Y-position $Y$, the roll angle $\phi$ and the yaw angle $\psi$. $X$, $Y$ and $\psi$ can be kinematically derived from the state variables:
\begin{align}
\psi(t+1)&=\psi(t)+r(t)\Delta t\label{eq4}\\
X(t+1)&=X(t)+\left\{v_{\rm x}(t)\cos\psi(t)-v_{\rm y}(t)\sin\psi(t)\right\}\Delta t\label{eq5}\\
Y(t+1)&=Y(t)+\left\{v_{\rm x}(t)\sin\psi(t)+v_{\rm y}(t)\cos\psi(t)\right\}\Delta t\label{eq6}
\end{align}

\subsection{Online Adaptation}
\label{3-B}
We use Continual-MAML \cite{c13} to help our agent adapt to new environments without forgetting past ones. Continual-MAML efficiently adapts to new environments while maintaining the ability to optimize starting parameters for previously seen environments. In our experiment, we refer to a change of road condition also as a change of task or environment.\par
In Continual-MAML, two kinds of parameters are used, meta parameters $\theta$ which tune to be optimized initial parameters and fast parameters $\tilde{\theta}$ which adapt to stationary local tasks. When the vehicle drives in a stationary task, the model adapts to the current task by updating fast parameters. When a task boundary is detected, meta parameters are adjusted using the accumulated task knowledge. After that, fast parameters are re-initialized using the meta parameters and they begin to adapt to the next stationary task. As a result, Continual-MAML allows the model to start updating from optimized initial values once the task boundary is detected and also adapt well to stationary tasks. This makes the model adaptation more efficient and robust to changes of environment.\par
Algorithm \ref{alg1} denotes the Continual-MAML flow in our scale-car environment which is slightly modified from the original algorithm. First, we will describe the original Continual-MAML algorithm and then explicitly state modifications to apply it to our application.\par
\begin{algorithm}[t]
    \caption{Continual-MAML for scale-car experiment}
    \label{alg1}
    \begin{algorithmic}[1]
        \REQUIRE $N_c$: the number of time steps used for one step
        \REQUIRE $\eta$, $l_r$: learning rate hyperparameter
        \STATE Initialize meta and fast parameters, $\theta$ and $\tilde{\theta}$
        \STATE Initialize buffer, $\mathcal{B}$ for incoming data
        \WHILE{}
        \STATE Sample ${\bm x}_t$ and ${\bm y}_t$ which consist of the past $N_c$ steps data
        \IF{Task boundary is detected}
        \STATE Sample training data ${\bm x}_{\rm train}$, ${\bm y}_{\rm train}\sim\mathcal{B}$
        \STATE Sample test data ${\bm x}_{\rm test}$, ${\bm y}_{\rm test}\sim\mathcal{B}$
        \STATE Fast adaptation $\theta'\leftarrow\theta-\eta\nabla_\theta\mathcal{L}(f_{\theta}({\bm x}_{\rm train}),{\bm y}_{\rm train})$
        \STATE Meta-Update $\theta\leftarrow\theta-l_r\nabla_\theta\mathcal{L}(f_{\theta'}({\bm x}_{\rm test}),{\bm y}_{\rm test})$
        \STATE Reset buffer $\mathcal{B}$
        \STATE Reset fast parameters $\tilde{\theta}_t\leftarrow\theta-\eta\nabla_{\theta}\mathcal{L}(f_{\theta}({\bm x}_t),{\bm y}_t)$
        \ELSE
        \STATE Add ${\bm x}_t$, ${\bm y_t}$ to buffer $\mathcal{B}$
        \STATE Fine-tuning $\tilde{\theta}_t\leftarrow\tilde{\theta}_{t-1}-\eta\nabla_{\tilde{\theta}}\mathcal{L}(f_{\tilde{\theta}_{t-1}}({\bm x}_t), {\bm y}_t)$
        \ENDIF
        \ENDWHILE
    \end{algorithmic}
\end{algorithm}
In the original algorithm, we need to initialize model parameters and a buffer (Alg. \ref{alg1}, L1-L2). Initial parameters are learned with MAML.
In the continual learning loop (Alg. \ref{alg1}, L3-L16) data used for model adaptation are sampled. If a task change is detected (Alg. \ref{alg1}, L5), meta parameters are updated using recent state measurements stored in the buffer (Alg. \ref{alg1}, L6-L9). After that, the buffer is cleared (Alg. \ref{alg1}, L10) and the fast parameters are reset to the meta parameters which are a more generally tuned set of initial parameters (Alg. \ref{alg1}, L11). Then the update process begins again.  If a task change is not detected, measurements are added to the buffer (Alg. \ref{alg1}, L13) and fast parameters are updated from their previous value (Alg. \ref{alg1}, L14).\par
Next, we concentrate on differences from the original algorithm. In our experiment, we don't use MAML in the pre-training phase and instead use the scheme referred to in Section \ref{3A} because we use data set in one road condition for pre-training. For the continual-loop, we define an update period $T_{\rm up}$ for model adaptation to a value that allows online adaptation to be carried out in real time, and data used for model adaptation are provided at each $T_{\rm up}$ period (Alg. \ref{alg1}, L4). These data consist of ${\bm x}$ and ${\bm y}$ from the past $N_c$ time steps. Although we would ideally cover 100 steps as we do in the pre-training phase, we needed to assign a smaller value $N_c$ to ensure real-time operation. The original algorithm assumes that task boundaries are unknown and introduces the mechanism to detect the boundaries. In our case, we assume that our agent knows the task boundaries and detects them correctly  (Alg. \ref{alg1}, L5)  in order to see the pure adaptation potential of Continual-MAML. Actually, visual information is effective for detecting changes and there are studies that utilize vision information for autonomous driving\cite{c26,c30,c31}. We leave incorporating the task detection algorithm for future work. If the task boundary is detected, meta parameters are updated as in the original algorithm. When updating the meta parameters, the original algorithm uses a mechanism to make a smooth behavior. However, in our experimental condition, this mechanism has little effect, so we don't adopt it to our algorithm at this time. When the task boundary is not detected, sampled data is added to the buffer (Alg. \ref{alg1}, L13). To reduce the computational burden, the buffer sizes for the training and test datasets are each set to one. In the original algorithm, sampled data are saved to the buffer each time and when the task boundary is detected or the buffer is full, it is cleared. With a buffer size of one, however, the data in either buffer will overlap with those contained in the other because each data point consists of multiple time step measurements. To avoid this overlap in training and test sets, in our experiment, we introduce a function to randomly decide whether to replace old data in a buffer and if so, which buffer's data to replace. In our algorithm, meta parameters are updated every 0.4 (s) while the original algorithm updates them only when the task boundary is detected or the buffer is filled. However, fast parameters are not initialized with meta parameters unlike when the task boundary is detected. In our experiment, the fast and meta parameters are pre-trained with data set in one road condition, and then, it is exposed to other road conditions in the continual learning phase. If the meta parameters haven't changed enough from their pre-trained values when the vehicle crosses a road condition transition, the meta parameters will be reinitialized to the pre-trained values. In order to address this issue, we periodically update meta parameters even though the vehicle has not crossed a task boundary. By doing it this way, the meta parameters are updated to be robust to existing conditions. We use Adam\cite{c32} and gradient descent for meta and fast parameter update, respectively, as performed in the original Continual-MAML paper\cite{c13}.
\subsection{Control Method}
We chose MPPI \cite{c14,c15,c16,c17,c18} as our baseline controller for all our models (fixed, gradient descent, and Continual-MAML). MPPI is a powerful nonlinear MPC based on the stochastic sampling approach which is able to address non-smooth and non-differentiable cost functions without approximations. In\cite{c14,c15}, the path integral control framework was improved so that both the mean and the variance of the sampling distribution could be tuned by a designer and it enabled the path integral control to converge quickly. They also compared the performance of MPPI to one of the most powerful nonlinear MPC formulations based on differential dynamic programming (DDP) and showed the superiority of MPPI in simulation. \cite{c16,c17,c18} deployed MPPI on a one-fifth scale car and showed the feasibility of MPPI near the limits of handling. \cite{c17,c18} utilized a neural network to model the vehicle dynamics.\par
As Graphics Processing Units (GPU) continue to develop further, MPPI will be able to compute on a larger number of samples, enabling faster and more accurate optimization. MPPI is an algorithm which has already been proven capable of stable planning and control near the limits of handling with scale cars and is expected to further improve performance in the future with more capable GPUs. For those reasons we chose MPPI described in\cite{c18} as our controller for showing the impact of modeling accuracy on overall control performance.\par
To accurately evaluate the impact of the dynamics model on the controller, we design a simple cost function to minimize with MPPI as follows:
\begin{equation}\label{eq7}
J=\alpha_1{\rm Track}({\bm x})+\alpha_2{\rm Speed}({\bm x})
\end{equation}
\noindent
where $\alpha_1$ and $\alpha_2$ are the coefficients of the track cost and the speed cost, respectively. Since it is known that using steering as well as braking is effective for collision avoidance\cite{c33}, tracking and speed are evaluated in the cost function. In addition, the cost function is designed to be simple to facilitate analysis of the impact of model accuracy on control. The details of the track cost and the speed cost are described in Section \ref{subsec:4-C}.
\section{EXPERIMENTAL SETUP}
\subsection{Experimental Environment}
\begin{figure}
\centering
\includegraphics[width=8cm]{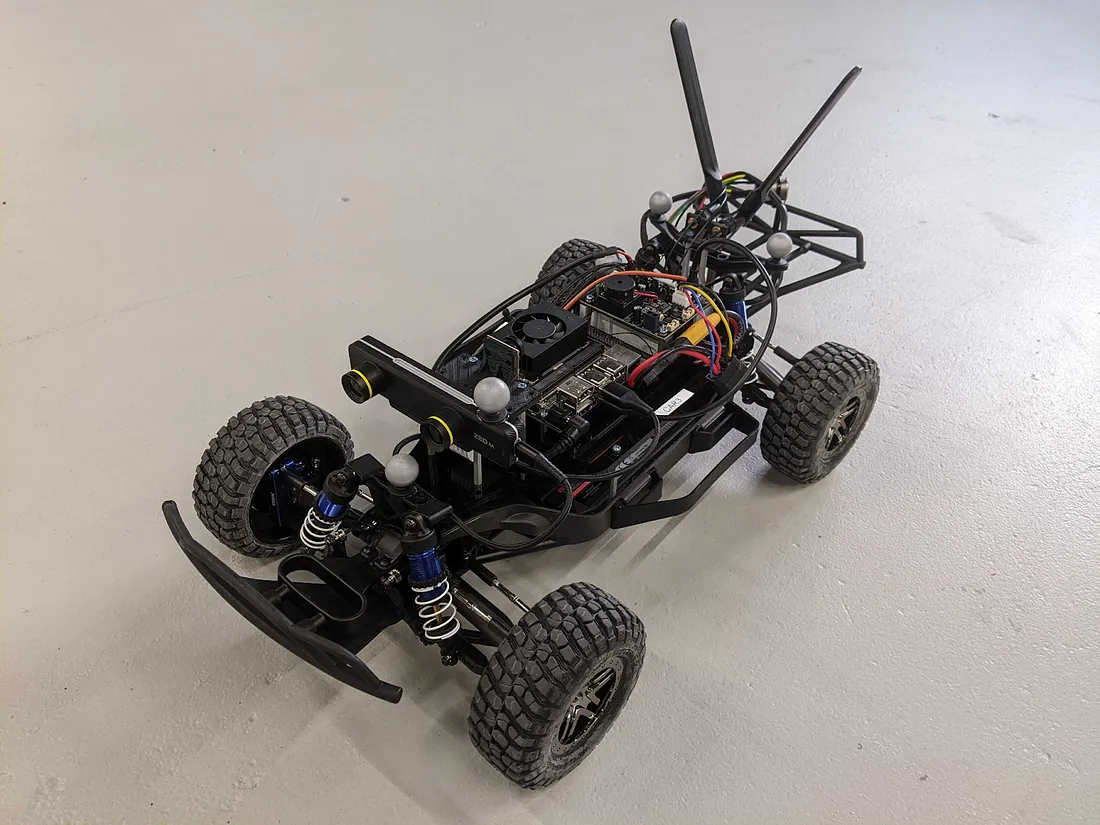}
\caption{TRIKart, a one-tenth scale-car which is used for our experiment. During experiments, electrical devices are covered by a plastic body and a bumper is mounted around the perimeter.}\label{fig1}
\end{figure}

\begin{figure}
\centering
\includegraphics[width=8cm]{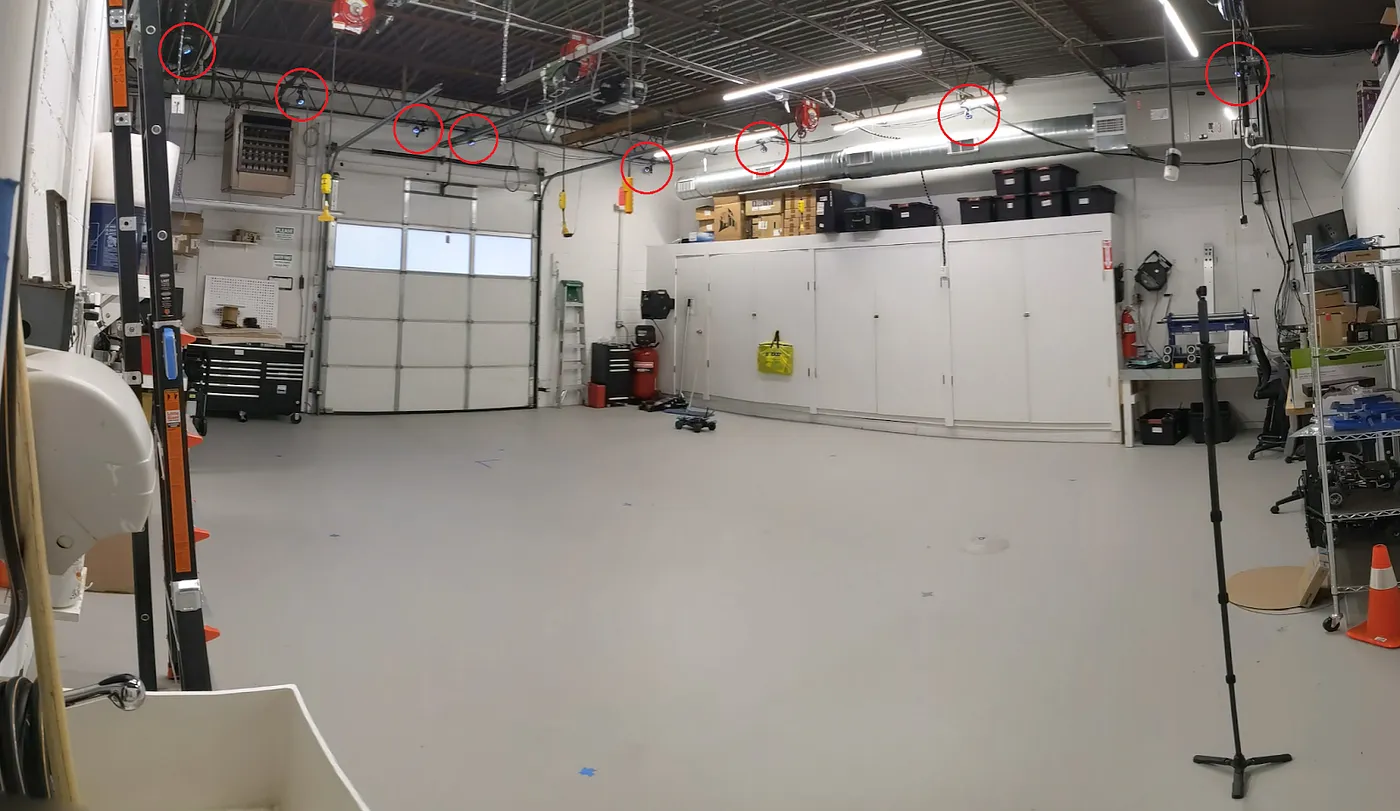}
\caption{Our test garage where 12 OptiTrack Prime 13 cameras are mounted on the ceiling. The cement floor in this garage is used for pre-training of a learned model.}\label{fig2}
\end{figure}

For the experiment, we utilize TRIKart \cite{c19}, a one-tenth scale-car platform shown in Fig. \ref{fig1}.
We use an Optitrack motion capture system consisting of 12 OptiTrack Prime 13 cameras as shown in Fig. \ref{fig2} to measure the vehicle's position and heading.
The system publishes these measurements at 120 Hz. From these values we derive velocity and acceleration. We use ROS to transmit information and run our algorithms.
\subsection{Model Setup}
\begin{figure}
\centering
\includegraphics[width=8cm]{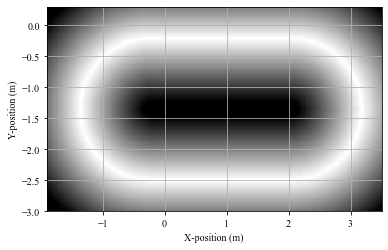}
\caption{The oval course used for our experiments. Colors represent a trajectory cost, that is, the cost of a white area is 0 and a black area is 1. The gray is linearly interpolated between 0 and 1.}\label{fig3}
\end{figure}

To evaluate the performance contribution of Continual-MAML, we prepare three models, a fixed model, an adaptive model using gradient descent and an adaptive model with Continual-MAML. For gradient descent, we just execute the update in Alg. \ref{alg1}, L14, while for Continual-MAML, the update process is as outlined by Alg. \ref{alg1}. For the experiment, the vehicle drives on a simple oval course to keep the problem complexity low and to provide a more straightforward analysis (Fig. \ref {fig3}).\par
All three models are pre-trained with the same data collected driving on a cement floor based on the learning scheme outlined in Section \ref{3A}. In addition, the pre-trained model was fine-tuned on the cement floor along the oval course with gradient descent. We did this to remove the possibility that the pre-trained model might not be well matched to the oval course; this way the difference in performance between the fixed model and the adaptive models derives most likely only from the handling of the road conditions, not also from adaptation to the course.\par
\begin{figure}
\centering
\includegraphics[width=8cm]{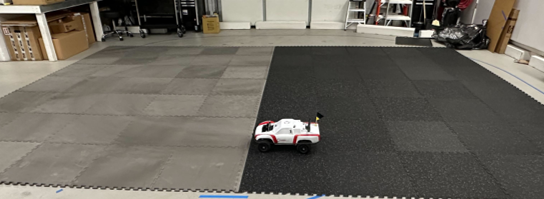}
\caption{Experimental road conditions for online adaptation. The right black mat is made with rubber which has high friction. The left gray mat is made with foam which has less friction than the rubber one.}\label{fig4}
\end{figure}

Fig. \ref {fig4} shows the experimental conditions used to evaluate the online adaptation models' performance on the road conditions not included in the training set.
Half of the course is on a rubber mat with higher friction than the cement floor. The other half of the road is on the foam back side of the rubber mat which has less  friction than the rubber mat. We used the backside of the rubber mat instead of a different surface in order to have a flat transition between the road conditions and avoid the change in road height as a possible experimental condition. The transition between road conditions occurs in the center of the straight part of the oval course. \cite{c13} introduces a mechanism to detect a task boundary. However, in our experiment we chose not to use that feature and instead provide the location of the road condition transitions to the algorithm.\par
Table \ref{table1} outlines outlines the parameter values used for the online adaptation algorithm. $T_{\rm up}$ is the update period of gradient descent and Continual-MAML. While a shorter duration is preferable, it is set to 80 (ms) to enable real-time execution. $N_c$ is the number of time steps of one step loss calculation. Ideally, this value should be set to 100, the same as during training. However, considering the computational load, it is set to 14. If this value is too small, the system becomes vulnerable to noise, resulting in decreased performance. $\eta$ and $l_r$ are learning rates for fast parameters and meta parameters, respectively. The buffer sizes for the training and test datasets are each set to one as mentioned in Section \ref{3-B} due to computational constraints.
\begin{table}
 \caption{Parameters of Online Adaptation} \label{table1}
 \centering
  \begin{tabular}{cr}
   \hline
   Item & \\
   \hline \hline
   $T_{\rm up}$ & 80 (ms) \\
   $N_c$ & 14 \\
   $\eta$ & 0.1 \\
   $l_r$ & $1.0\times 10^{-4}$ \\
   Buffer size & 1 training buffer and 1 test buffer\\
   \hline
  \end{tabular}
\end{table}

\subsection{Control Setup}
\label{subsec:4-C}
Here, we provide the details of (\ref{eq7}) in our test environment. The track cost is defined in a costmap as illustrated in Fig. \ref {fig3}. In Fig. \ref {fig3}, the cost value of a white area is 0 and the cost of a black area is 1. The gray area between the white and black areas is linearly interpolated between 0 and 1 along the track normals.\par
The speed cost is defined as follows:
\begin{equation}\label{eq8}
{\rm Speed}({\bm x})=(v_{\rm x}-v_{\rm ref})^2
\end{equation}
\noindent
where $v_{\rm ref}$ is the reference velocity and is kept constant in our experiments. $v_{\rm x}$ is the longitudinal velocity of the vehicle which is predicted by the vehicle dynamics model. MPPI provides the steering angle and acceleration command sequence which optimizes the sum of the cost function over 100 prediction steps.\par
For the experiment, we set $\alpha_1=600$, $\alpha_2=25$ and $v_{\rm ref}=3.2$ (m/s) in (\ref{eq7}).

\section{RESULTS}
We probe the performance of the learned dynamics model from two main angles. As a learning problem, we first demonstrate the effectiveness of our approach in learning the correct dynamics coefficients using inference results. We then probe the performance of the resultant model when combined with a MPPI controller.
\subsection{Inference}
We evaluate the model accuracy using inference performance. To compare the inference performance, the vehicle autonomously drives clockwise on the mats (Fig. \ref{fig4}) using a fixed learned dynamics model while recording the vehicle state. After that, the inference error was calculated by sequentially giving the measured vehicle state as the input to each of the three models, namely the fixed model, the adaptive model with gradient descent and the adaptive model with Continual-MAML. We evaluate performance using the cumulative loss over $N_c$ steps, which is minimized during online adaptation.\par
\begin{figure}
        \centering
        \includegraphics[width=8cm]{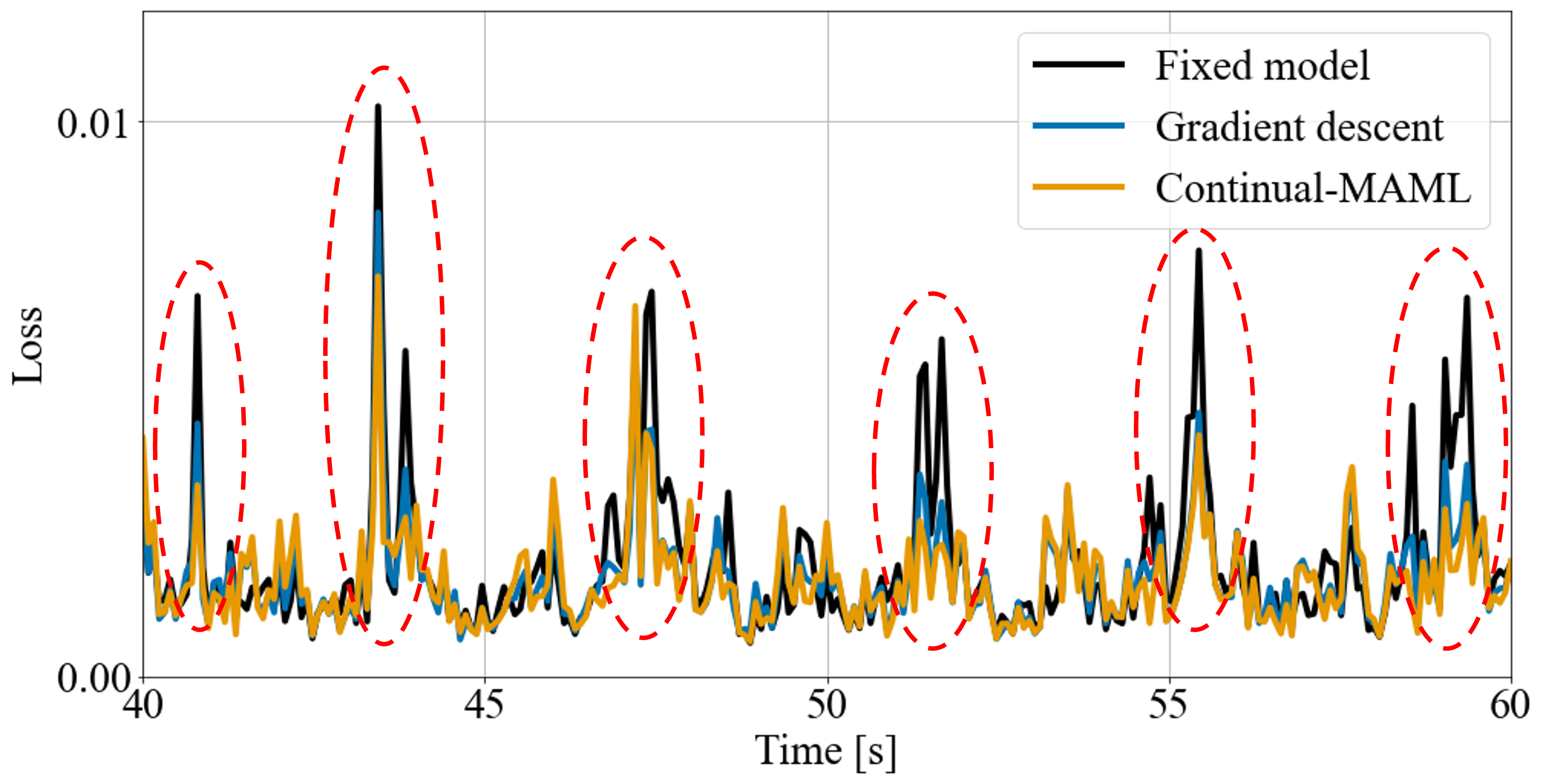}
        \caption{Enlarged view of inference loss for each model between 40 (s) to 60 (s). Each peak characterized by red dotted ellipses corresponds to turning on rubber mat. Especially in this region, differences between each method are significant. Gradient descent outperforms the fixed model and it shows effectiveness of adaptation. Continual-MAML outperforms gradient descent especially in turning on rubber mat and it implies Continual-MAML achieves quick and effective adaptation by starting update from optimal initial parameters when a road change is detected.}\label{fig5}
\end{figure}
Fig. \ref{fig5} is an enlarged view of the inference loss for each model between 40 (s) to 60 (s). Since the first 40 (s) have almost the same performance difference as in Fig. \ref{fig5}, we show the enlarged view to make it easier to see differences. As illustrated in Fig. \ref{fig5}, the adaptive models significantly outperform the fixed model in terms of loss value. The periodical peaks of loss in Fig. \ref{fig5} correspond to cornering on the rubber mat, and Continual-MAML shows better inference performance than gradient descent particularly during those maneuvers. This implies that Continual-MAML can efficiently learn adequate parameters by starting its update cycle from well tuned initial parameters.\par
\begin{table}
        \caption{Cumulative Mean Square Inference Loss over 60 (s)}\label{table2}
        \centering
        \begin{tabular}{crrr}
        \hline
        Test No. & Fixed model & Gradient descent & Continual-MAML \\
        \hline \hline
        $ N_1 $ & $2.56\times10^{-3}$ & $2.10\times10^{-3}$ &$\mathbf{1.98\times10^{-3}}$ \\
        $ N_2 $ & $2.63\times10^{-3}$ & $2.20\times10^{-3}$ &$\mathbf{2.16\times10^{-3}}$ \\
        $ N_3 $ & $2.75\times10^{-3}$ & $2.29\times10^{-3}$ &$\mathbf{2.23\times10^{-3}}$ \\
        \hline
        \end{tabular}
\end{table}
Table \ref{table2} shows the cumulative mean square error loss over 60 (s). We performed 3 iterations of the test and from the results in Table \ref{table2}, we can see that Continual-MAML achieved the best inference performance for the same driving data.
\subsection{Control}
To evaluate the contribution of the dynamics model accuracy to the control performance, we deploy the MPPI algorithm with the cost function described in (\ref{eq7}) using each of the three dynamics models on a TRIKart platform. For the adaptive models, model parameters are updated in real-time with gradient descent and Continual-MAML, respectively. The vehicle drives several laps clockwise for each model.\par
\begin{figure}
        \centering
        \includegraphics[width=8cm]{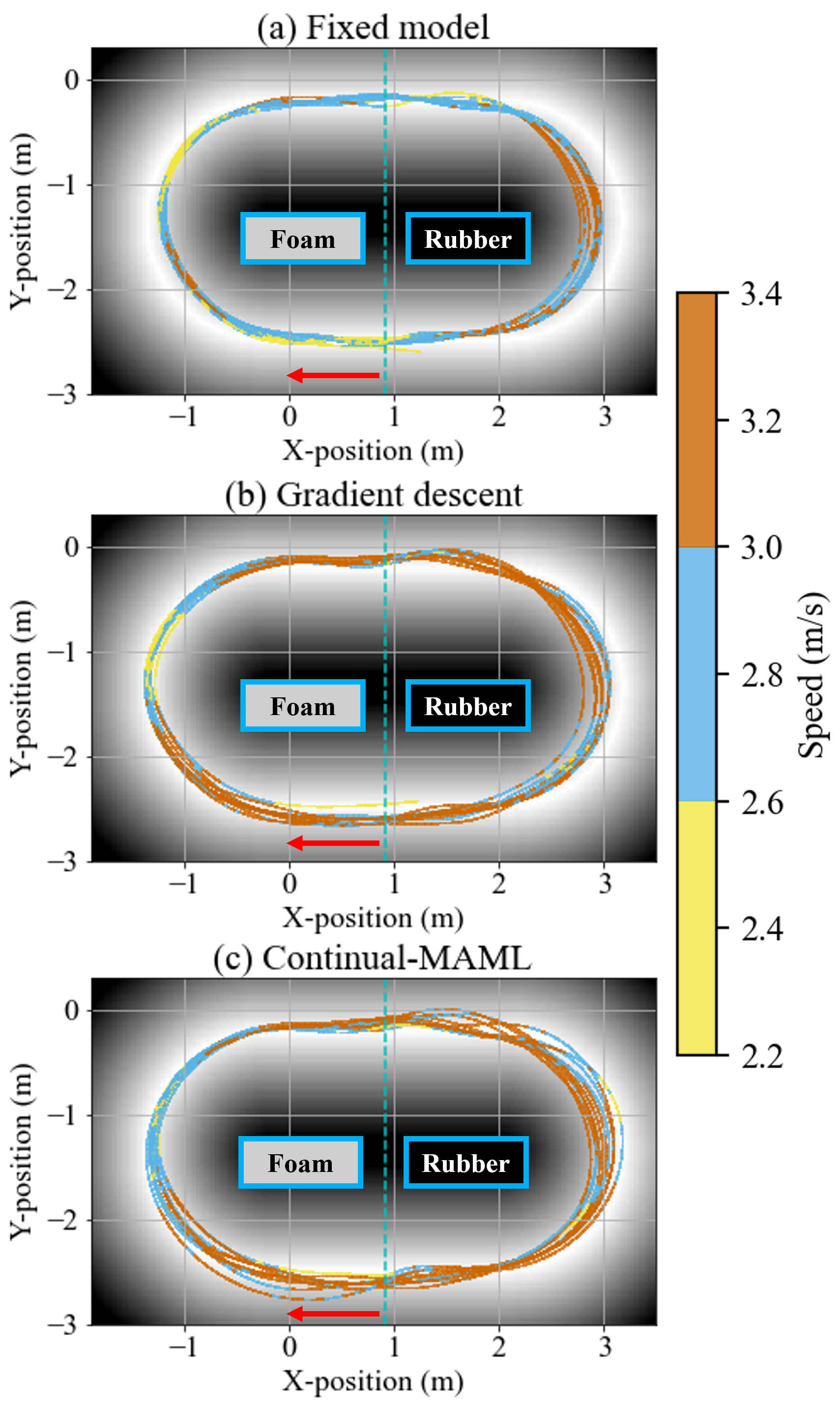}
        \caption{Clockwise vehicle trajectories for 10 laps controlled using MPPI with a target velocity, 3.2 (m/s). MPPI optimizes based on (a) a fixed model, (b) an adaptive model with gradient descent and (c) an adaptive model with continual-MAML. Gradient descent and Continual-MAML trajectories obviously follow the target path indicated by the white ranges better than the fixed model. Speed profile with Continual-MAML is the closest to the target velocity.}\label{fig6}
\end{figure}

Fig. \ref{fig6} illustrates the first ten lap trajectories with each model. At the top of Fig. \ref{fig6}, the plot shows that the fixed model went off course on the inside of the corner, especially on the rubber mat. We posit that this is because the friction of rubber mat is much higher than that of the cement floor where the training data for the fixed model was collected causing more reaction to steering input than expected. The lower two plots show that the two adaptive models  improved their trajectory error by learning better model parameters on mats in real-time. Moreover, the velocity of the adaptive models was closer to the target value than that of the fixed model. The velocity of the adaptive model using gradient descent was slower in the latter half of the corner on the side of the course with foam mats. We hypothesize that the model learned an acceptable speed on the rubber road surface and when entering the foam road surface, the model started learning from the model parameters adapted to the rubber surface. As a result, the model initially ran at a speed appropriate for the rubber surface, but too fast for the foam surface. The controller then had to slow down to avoid deviating from the target path. Alternatively, the adaptive model with Continual-MAML started learning from initial parameters adapted to both new surfaces when the vehicle entered a new road condition. Therefore, Continual-MAML was able to keep a higher velocity on both road surfaces.\par
Table \ref{table3} shows the statistics for the control error evaluated as the sum of the tracking and velocity error (\ref{eq7}). Ten tests were executed for each model. Each test consisted of 18 laps of the course. Table \ref{table3} summarizes the average value of the control error for each test, and average error is evaluated over the second to the eighteenth lap. As shown in Table \ref{table3}, MPPI with Continual-MAML has a lower control error than with gradient descent.
\begin{table}
        \caption{Control Error Statistics Consisting of Trajectory and Speed Error}\label{table3}
        \centering
        \begin{tabular}{crrr}
        \hline
        Test No. & Fixed model & Gradient descent & Continual-MAML \\
        \hline \hline
        $ N_1 $ & $58.44$ & $49.70$ &$46.68$ \\
        $ N_2 $ & $61.20$ & $53.67$ &$44.18$ \\
        $ N_3 $ & $61.76$ & $46.34$ &$44.47$ \\
        $ N_4 $ & $60.69$ & $51.06$ &$45.90$ \\
        $ N_5 $ & $55.57$ & $50.70$ &$46.02$ \\
        $ N_6 $ & $57.62$ & $51.49$ &$42.72$ \\
        $ N_7 $ & $59.87$ & $47.03$ &$46.12$ \\
        $ N_8 $ & $59.16$ & $46.64$ &$40.96$ \\
        $ N_9 $ & $56.44$ & $44.66$ &$42.85$ \\
        $ N_{10} $ & $59.18$ & $44.41$ &$41.21$ \\
        \hline
        Mean & $58.99$ & $48.57$ &$\mathbf{44.11}$ \\
        Min. & $55.57$ & $44.41$ &$\mathbf{40.96}$ \\
        \hline
        \end{tabular}
\end{table}
\section{CONCLUSIONS}
In this paper, we apply a state of the art Continual-MAML algorithm to the online setting of the parameters of a learned vehicle dynamics model. We evaluate adaptation performance of it using the TRIKart scale car platform. We utilize a neural network to capture the vehicle dynamics behavior and its weights and biases are pre-trained on data collected by driving on a cement floor. We probe the online dynamics learning from two main angles. The first is its inference performance and the second is its contribution to the speed and tracking performance of a model-based autonomous controller. For both evaluations, we drove the scale-car autonomously on an oval course half of whose surface consisted of rubber mats with the other half consisting of foam mats, surfaces which have not been in the training set. We utilized MPPI as our baseline controller.\par
For inference evaluation, we measured the loss for each of our candidate models. Online adaptation aims to minimize the inference loss, and so consequently both the model using gradient descent and the one using Continual-MAML outperformed the fixed model. However, Continual-MAML outperformed both the fixed model and gradient descent, especially during cornering on the rubber mat and this implies that Continual-MAML can quickly and efficiently learn adequate parameters by starting its update cycle from well optimized initial parameters.\par
For the models' contribution to the performance of autonomous driving with model-based control, we evaluated the resultant cost value consisting of track and speed errors. With Continual-MAML, the mean cost value was improved 25\% compared to the fixed model and 9\% compared to the model using gradient descent. In addition, Continual-MAML enabled the controller to improve minimum cost value 26\% compared to the fixed model and 8\% compared to the gradient descent.\par
Future work includes the introduction of automatic detection algorithms for a task boundary in autonomous driving scenarios and the extension of our approach to more challenging environments that may actually occur in the real world, such as slippery roads.
\addtolength{\textheight}{-12cm}   








\end{document}